%% file: acl_latex.tex
\pdfoutput=1

\documentclass[11pt]{article}

\usepackage[preprint]{acl}

\usepackage{times}
\usepackage{latexsym}

\usepackage[T1]{fontenc}

\usepackage[utf8]{inputenc}

\usepackage{microtype}

\usepackage{inconsolata}
\usepackage{graphicx}
\usepackage{booktabs}
\usepackage{diagbox}
\usepackage{todonotes}
\usepackage{multirow}

\title{Beyond Memorization: The Challenge of Random Memory Access\\in Language Models}

\author{Tongyao Zhu$^{1,2}$\quad Qian Liu$^{1}$\thanks{Corresponding authors.}\quad Liang Pang$^{3*}$\quad Zhengbao Jiang$^{4}$ \\
\bf{Min-Yen Kan$^{2}$\quad Min Lin$^{1}$} \\
$^1$Sea AI Lab \quad $^2$National University of Singapore\\
$^3$Institute of Computing Technology, CAS\quad
$^4$Carnegie Mellon University \\
\texttt{tongyao.zhu@u.nus.edu}\quad\texttt{\{liuqian,linmin\}@sea.com} \\
\texttt{pangliang@ict.ac.cn}\quad \texttt{zhengbaj@cs.cmu.edu}\quad \texttt{knmnyn@nus.edu.sg} \\   
}

\begin{document}
\maketitle
\begin{abstract}
Recent developments in Language Models (LMs) have shown their effectiveness in NLP tasks, particularly in knowledge-intensive tasks.
However, the mechanisms underlying knowledge storage and memory access within their parameters remain elusive.
In this paper, we investigate whether a generative LM (e.g., GPT-2) is able to access its memory sequentially or randomly.
Through carefully-designed synthetic tasks, covering the scenarios of full recitation, selective recitation and grounded question answering, we reveal that LMs manage to sequentially access their memory while encountering challenges in randomly accessing memorized content. We find that techniques including recitation and permutation improve the random memory access capability of LMs. 
Furthermore, by applying this intervention to realistic scenarios of open-domain question answering, we validate that enhancing random access by recitation leads to notable improvements in question answering. The code to reproduce our experiments can be found at \url{https://github.com/sail-sg/lm-random-memory-access}.
\end{abstract}

\input{intro}
\input{related-work}
\input{seq-rand-challenge}

\input{solution}
\input{case-study-odqa}

\input{conclusion}

\input{limitation-and-ethics}

\bibliography{anthology,custom}
\appendix
\input{appendix}

\end{document}

%% file: intro.tex
\section{Introduction}

Language models (LMs) have recently showcased outstanding abilities in NLP tasks with a large amount of memory stored in their parameters \citep{GPT3-NIPS, ouyang2022training}. Through pre-training on large text corpora, LMs memorize factual knowledge about the world \citep{zhou2023lima}. Consequently, they show great performance in knowledge-intensive tasks \citep{petroni-etal-2021-kilt} such as open-domain question answering \citep{kamalloo-etal-2023-evaluating, ziems-etal-2023-large, mallen-etal-2023-trust}. There is a growing interest in considering LMs as knowledge bases \citep{wang-etal-2021-generative, heinzerling-inui-2021-language,petroni-etal-2019-language, cao-etal-2021-knowledgeable, AlKhamissi2022ARO}. Despite the recent advances in applying LMs to solve downstream tasks, the fundamentals of how LMs store knowledge and access memory in their parameters remain a subject of ongoing research and intrigue \citep{Tirumala2022MemorizationWO, Zhu2023PhysicsOL, AllenZhu2023Physics3.2, berglund2023reversal}.

\begin{figure}[tb]
  \centering
    \includegraphics[width=1\columnwidth]{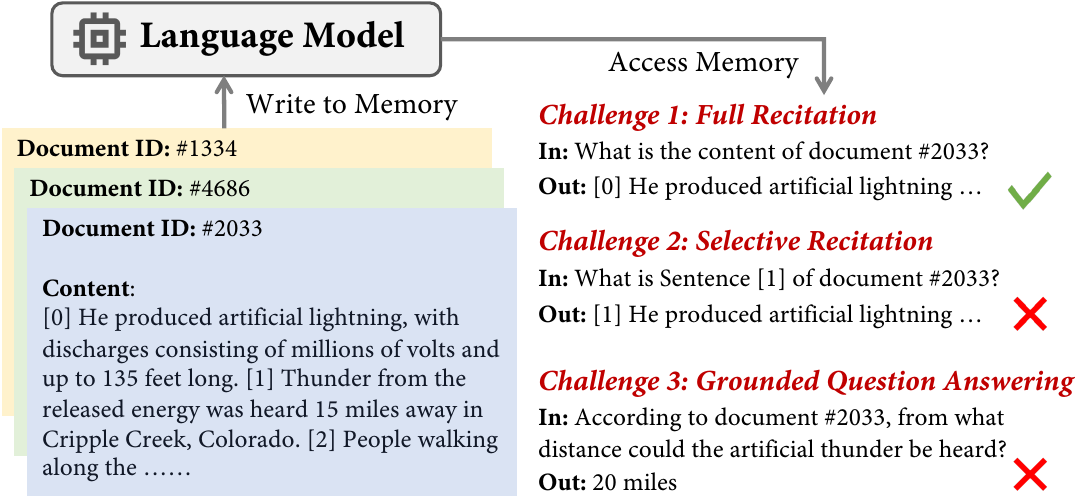}
  \caption{An illustration of our investigation of memory access pattern in language models. We find that the model accesses its parametric memory largely in a sequential manner, and faces difficulty in randomly accessing the content in the middle of memorized strings.
  }
  \label{fig:seqvsrand}
\end{figure}

In this paper, we draw inspiration from memory-accessing patterns observed in computer systems to explore whether LMs can access their parametric memory in a sequential or random manner. 
We extrapolate these concepts to investigate LMs and delineate two memory access patterns: \textit{sequential memory access} means that the model starts from the beginning of a memorized sequence, progressing through the content in consecutive order. Conversely, \textit{random memory access} denotes that the model can commence from any location within the memorized content, without needing to start from the beginning. For instance, reciting a memorized poem line by line is considered sequential access, while directly starting from the third line involves random access.  

With these concepts, we design experiments with both synthetic and real data to evaluate the language model's ability to perform sequential or random access to memorized content, as illustrated in Figure \ref{fig:seqvsrand}. 
We limit our study to 
decoder-only language models because of their increasing popularity and capability \citep{radford2019language,GPT3-NIPS, touvron2023llama, touvron2023llama2, Jiang2023Mistral7}. We first ask the model to memorize key--value pairs of various types and show that the model is able to sequentially read memorized content to a satisfying degree. Next, we test the model's random access ability by training it to recite a sentence or find an answer to a question in a memorized passage. In such tasks, the model's performance falls drastically when it is required to extract a span in the middle of a passage, revealing its incapacity to randomly access its memory. 

Given that language models struggle to perform random access to their memory, we pursue two means for mitigation: \textit{recitation} at inference time, and \textit{permutation} during training. Recitation enables the model to sequentially read its parametric memory first before performing a task. The model's performance can thus be enhanced by utilizing the recited content in its context window. We also show that simply permuting sentences in a passage during training to memorize content also improves performance.

We finally verify the challenge of random access through a case study on open-domain question answering. We reduce the difficulty of the task by allowing the model to memorize passages with ground-truth answers, yet we find that the model benefits the most from such memorization when it is allowed to recite a relevant passage and then answer the question. Overall, we make several contributions to further understand the memory access mechanisms of decoder-only language models: 
\begin{itemize}
    \item We show that language models can access their memory sequentially and can reproduce memorized content,  but encounter significant challenges in random memory access.
    \item  We find solutions to mitigate the challenge of random access by permuting memorized content or explicitly reciting the memory before performing tasks. 
    \item We demonstrate the effect of poor random memory access ability in open-domain question answering, showing that the challenge could have broader implications on the applications of language models.
\end{itemize}

%% file: related-work.tex
\section{Related Work}

\paragraph{Memorization in Language Models.}  Large language models store a considerable amount of knowledge in their parameters \citep{petroni-etal-2019-language,heinzerling-inui-2021-language}. They memorize useful knowledge such as facts and commonsense~\citep{zhao2023large}, but also sensitive personal information such as emails or phone numbers \citep{Carlini2020ExtractingTD, huang-etal-2022-large}. Existing approaches to understanding memorization include fine-grained analysis to locate the neuron that is associated with the knowledge~\citep{NEURIPS2022_6f1d43d5,liu2024the} or macro analysis to understand the overall dynamics of memorization \citep{Tirumala2022MemorizationWO,speicher2024understanding}. In this study, we do not aim to analyze the mechanisms of writing to language model's memory. Instead, we consider the language model as a black-box memory store and focus mainly on how the model accesses its memory.

\paragraph{Knowledge Injection.}

Our investigation requires writing new content to the model's parametric memory. There are mainly two ways to perform such knowledge injection without changing the model architecture \citep{ovadia2024finetuning,balaguer2024rag}: fine-tuning or retrieval augmentation. Retrieval augmentation \citep{NEURIPS2020_6b493230,shi2023replug} retrieves relevant information and puts it into the model's context while fine-tuning directly updates the model parameters. As the goal of our study is to investigate how the model accesses its parametric memory after writing to the memory, we choose finetuning as the method for introducing new knowledge to the model. 

\paragraph{Knowledge Retrieval.}
Previous works have shown that using prompts can effectively retrieve knowledge stored in large language models~\citep{Bouraoui2019InducingRK, jiang-etal-2021-know, wang-etal-2021-generative}. We follow earlier work to use prompts to query the model to access and regenerate memorized content. However, a notable difference is that prior work focuses on finding optimised methods to elicit the model's knowledge obtained during pretraining \citep{youssef-etal-2023-give, 10.1145/3560815, yu2023generate}, while we directly use unique keys for memorizing and retrieving content.

\paragraph{Language Model as a Document Index.}
We consider the language model as a memory store for passages, which is related to the recent advances in adopting a language model as an index for document storage and retrieval \citep{10.1145/3476415.3476428, tay2022transformer, wang2023neural, Zeng2023ScalableAE}. In such indexes, each document is associated with a document identifier (ID), which could be keywords \citep{ren-etal-2023-tome,SEAL_NEURIPS2022_cd88d62a,lee-etal-2023-glen,lee-etal-2023-nonparametric} or numbers \citep{tay2022transformer, wang2023neural,zhuang2022bridging,Zhou2022UltronAU}. We also follow the practice and assign an ID to each document for storing and retrieving the documents. However, we do not ask the model to retrieve a relevant ID to a question. Instead, we provide the ID in the input, and investigate the possibility of sequentially or randomly accessing the corresponding document content. 

%% file: seq-rand-challenge.tex
\section{Investigating Sequential and Random Memory Access}

In this section, we investigate the ability of a language model to sequentially or randomly access its memory stored in the parameters. First, we provide formulations of language models serving as a memory bank of passages (\S\ref{sec:task-formulation}). Within this framework, we define \textit{sequential memory access} as the process of starting from the beginning of a memorized passage and progressively generating subsequent content. In contrast, we conceptualize \textit{random memory access} as the model’s ability to initiate recall from any chosen location in a memorized passage and accurately regenerate the subsequent content. Based on these definitions, we first investigate the model's sequential memory access ability by requiring it to recite full passages word by word (\S\ref{section:sequential-access}). Next, we test the random memory access ability of the model by asking it to recite selected sentences from memorized passages (\S\ref{section:selective-recitation}). We further assess the model's random access proficiency through a more challenging task involving question answering (\S\ref{section:grounded-qa}).

\subsection{Task Formulation} \label{sec:task-formulation}

We abstract the language model as a memory bank and investigate its sequential or random access ability. We adopt a simple definition of a memory bank as a key--value store $\mathcal{D}=\{k_i:p_i\}$, where $k_i$ represents a unique identifier (ID) assigned to the content of the $i$-th passage\footnote{We use ``document'' and ``passage'' interchangeably to refer to a chunk of text.}.

There are two core functions that a memory bank needs to support: \textit{reading} and \textit{writing}. Given that our memory bank is embodied as a language model, it is not straightforward to write and read the model's memory.
Following previous work~\citep{Zhu2023PhysicsOL,wang-etal-2021-generative}, for \textit{writing} to the memory bank, we use fine-tuning to update the model's parameters.
For \textit{reading}, we use prompting to elicit the model’s memory.
Specifically, for each passage $p_i$ with its corresponding identifier $k_i$, we create two types of data instances: \textbf{writing}, $S_{write}(k_i, p_i)$ and \textbf{reading}, $S_{read}(k_i) \rightarrow p_i$, where $S_{write}$ and $S_{read}$ denote the prompts detailed in Appendix \ref{appendix:prompt-for-memorization}.

As the primary goal of our study is to test whether the model can read (access) its stored content sequentially or randomly, we vary the \textit{reading} function across different experiments. Given a corpus consisting of $M$ passages, we split the corpus into two subsets: $T$ training passages and $V=M-T$ validation passages. We adopt a mixed training strategy as described by \citet{Zhu2023PhysicsOL}: During the training stage, we include $S_{write}$ and $S_{read}$ instances of \textit{T} training passages, as well as $S_{write}$ instances of \textit{V} validation passages. Our objective is for the model to learn to associate each identifier with its passage content by training on the reading and writing instances of the training passages. During evaluation, we prompt the model with the $S_{read}$ instances of the \textit{V} validation passages to test the model's memory access pattern.

\subsection{Sequential Access: Full Recitation}
\label{section:sequential-access}

We test the sequential access ability of the language model by asking it to reproduce the full passage content. Specifically, given an ID, the model is prompted to start from the beginning of the corresponding memorized passage and generate tokens consecutively. We evaluate the model’s performance to reproduce the content on the $V$ validation passages, which requires the model to both memorize the passage content and sequentially access the memory with the provided key.

\paragraph{Setup.}\label{sec:memorize_setup}
To investigate whether the model can handle identifiers and passage content of different types, we set $T=400$ and $V=40$ and consider the following variations. For the type of passage content $p$, we examine two categories: (1) natural language (\textit{NL}), comprising Wikipedia paragraphs from SQuAD \citep{rajpurkar-etal-2016-squad}, and (2) random strings (\textit{Rand}), where each \textit{NL} passage is substituted with a space-separated alphanumeric string maintaining the same number of tokens. Regarding the type of $k$ (i.e., passage IDs), we explore three forms: (1) numerical strings (\textit{Num}), such as `\#123'; (2) rare random tokens (\textit{Rare}), adopting the approach of \citet{Ruiz2022DreamBoothFT} by random sampling three infrequent tokens; (3) article title (\textit{Title}) of the Wikipedia page to which the passage belongs.

    We adopt the GPT2-large model~\citep{radford2019language} with 774M parameters as the base model. For better string memorization ability~\citep{Stevens2023MemorizationFG}, we use a pretrained checkpoint\footnote{\url{https://huggingface.co/gpt2}} instead of training the model from scratch. We fine-tune the model for 100 epochs to ensure that the model fully converges, with a learning rate of $3\times10^{-5}$. We measure memorization using both the BLEU score \citep{papineni-etal-2002-bleu} and the Exact Match (EM) score, indicating the similarity between the generated content and the ground-truth passage.

\begin{table}[tb]
    \centering
    \begin{tabular}{lccc}
    \toprule
   & \textbf{Title (ID)} & \textbf{Num (ID)} & \textbf{Rare (ID)} \\ \midrule
    psg=\textit{NL} & 96.2\,/\,85.0 & 96.7\,/\,95.0 & 73.4\,/\,72.5 \\
    psg=\textit{Rand} & 96.7\,/\,95.0 & 96.7\,/\,95.0 & 96.7\,/\,95.0 \\ \bottomrule
    \end{tabular}
    \caption{BLEU\,/\,Exact Match scores of reading from memory with different types of IDs and passage content. }
    \label{tab:find_passage_diff_type}
\end{table}

\paragraph{Discussion.} Table \ref{tab:find_passage_diff_type} shows that the model is able to sequentially access memorized content, with high BLEU and EM on validation passages. The model's sequential access capability is further demonstrated by its adaptability to varying types of IDs and passages. Specifically, using titles or numbers as keys for natural language passages achieves higher performance than using rare tokens. We suspect that models might have difficulty associating rare tokens with the natural language content. Remarkably, the model's access ability extends to passages composed of random characters ($Rand$).

\begin{figure}[tb]
  \centering
    \includegraphics[width=\columnwidth]{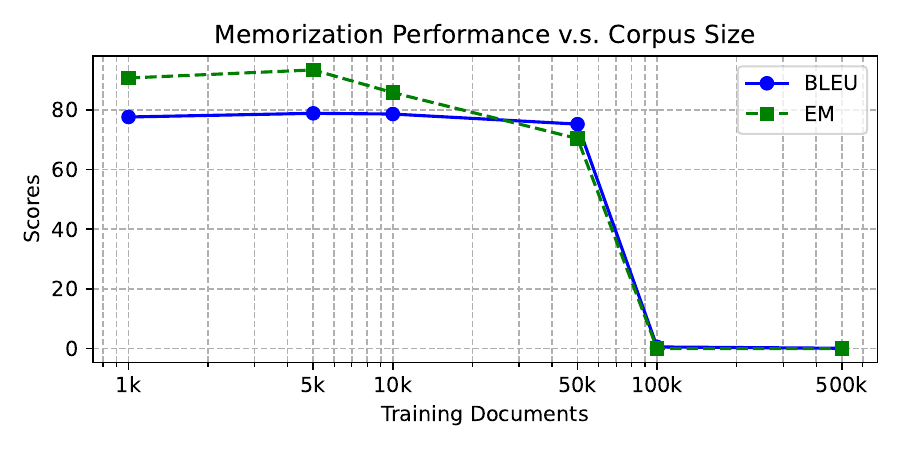}
   \caption{EM and BLEU for reading validation passages, varying the number of training passages.
    We calculate EM using only the first 25 tokens, as the model tends to continue generation beyond the max passage length (25).} 
  \label{fig:memlimit}
\end{figure}

To further test the memory capacity of the model, we carry out an additional experiment where we set the passage type to $Rand$ and identifier type to $Rare$ and construct passages each with 25 random tokens. As illustrated in Figure~\ref{fig:memlimit}, we fix $V$ as 1k and increase $T$ gradually from 1k to 500k to examine the ability of sequential memory access.

We observe that even with a training passage count of 50k, the GPT2-large model accurately reproduces over 70\% of memorized validation passages.
However, there is also a bottleneck in parametric memory: the performance drops to nearly zero when the passage count exceeds 100k.
We attribute this bottleneck to the difficulty in training, as the model fails to converge on memorizing all the passages.
Therefore, in subsequent experiments, we carefully manage the corpus size to ensure that the model memorizes all passages.

\subsection{Random Access: Selective Recitation}\label{section:selective-recitation}
 
Selective recitation is a straightforward synthetic task: asking the language model to reproduce a specific sentence of a memorized passage. This task is designed for its simplicity, as it does not require the model’s understanding of passage content. The focus is solely on the model’s capacity to access segments in a memorized passage. Successful random access would be indicated by the model’s ability to reproduce any sentence from within memorized passages, regardless of position. 

 \paragraph{Setup.}
We follow \citet{Mallick2023AdaptingPG} to place markers at the boundaries of each sentence, obtained by the NLTK sentence splitter\footnote{\url{https://www.nltk.org/api/nltk.tokenize.sent\_tokenize.html}}: a passage is formatted as ``\textit{[0] {sent0} [0] [1] {sent1} [1], ...,}''. In this case, the model only needs to learn to copy the content between these markers. Our selective recitation task requires the model to recite the $j$-th sentence of passage $p_i$ based on the given passage ID $k_i$. The reading function is now $S_{read}(k_i, j) \rightarrow p_i[j]$, such as ``\textit{What is sentence [1] of Document \#2033?}'' shown in Figure~\ref{fig:seqvsrand}. For reference, we also test the model's performance in a baseline where the passage content is provided in the context window. 

As we are testing for exact memorization, we use BLEU and EM scores to evaluate the model. Similar to \S\ref{sec:memorize_setup}, we use $T=400$ training and $V=40$ validation passages, with 1994 sentences and 200 sentences respectively. We set the type of ID to be $Title$ and only include passages with more than 3 sentences. All other hyperparameters stay the same as \S\ref{sec:memorize_setup}.

\begin{figure}[tb]
  \centering
    \includegraphics[width=1.0\columnwidth]{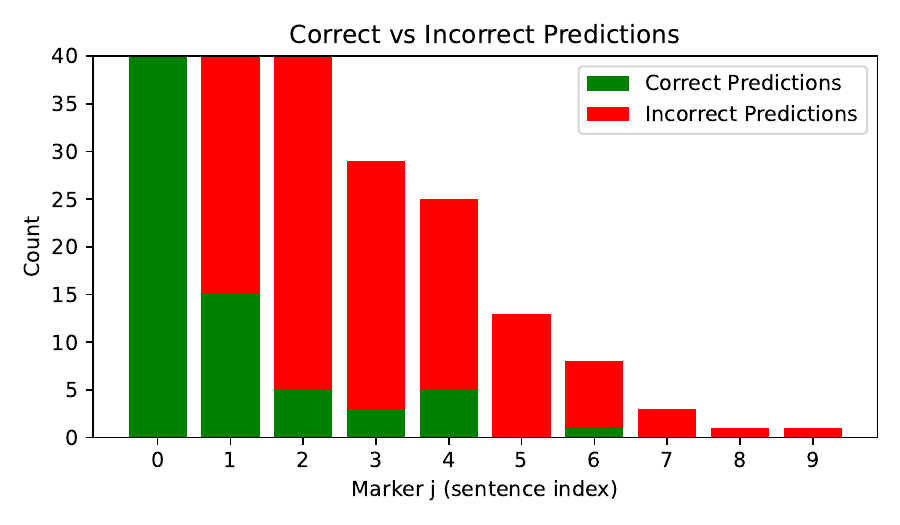}
  \caption{A stacked bar plot showing the accuracy of ID-guided sentence recitation with different marker numbers. The performance decreases significantly as the sentence index grows, revealing the model's incapability in accessing middle sentences. }
  \label{fig:accuracy-vs-marker}
\end{figure}

\begin{table*}[tb] 
    \centering
\begin{tabular}{lccccccc}\toprule
 &\multicolumn{2}{c}{\textbf{Title (ID Type)}} &\multicolumn{2}{c}{ \textbf{Rare (ID Type)}} &\multicolumn{2}{c}{\textbf{Num (ID Type)}} \\
 \cmidrule(lr){2-3} \cmidrule(lr){4-5} \cmidrule(lr){6-7}
 \textbf{Setup} &EM &F1 &EM &F1 &EM &F1 \\\midrule
 \multicolumn{7}{c}{\textit{w/o passage memorization}} \\
  Closed-Book QA (lower bound) &9.0 &16.6 &9.0 &16.6 &9.0 &16.6 \\
  Open-Book QA (upper bound) &73.7 &79.3 &73.7 &79.3 &73.7 &79.3 \\
\midrule
 \multicolumn{7}{c}{\textit{w/ passage memorization}} \\
  Grounded QA \textit{w/ Golden ID} &26.7 &35.6 &20.7 &28.7 &24.3 &32.6 \\
   ~~~~~~~$\hookrightarrow +$ Recitation &\textbf{59.7} &\textbf{68.0} &\textbf{54.7} &\textbf{62.1} &\textbf{57.7} &\textbf{66.2 }\\
 Grounded QA \textit{w/ Random ID} &20.7 &28.9 &20.7 &28.3 &23.3 &31.6 \\
  ~~~~~~~$\hookrightarrow +$ Recitation &16.0 &20.4 &18.7 &23.6 &18.7 &23.1 \\
  Grounded QA \textit{w/o ID} &22.0 &31.0 &22.0 &31.0 & 22.0 &31.0 \\
  ~~~~~~~$\hookrightarrow +$ Recitation &26.3 &33.1 &26.3 &33.1 &26.3 &33.1 \\

\bottomrule
\end{tabular}
    \caption{EM and F1 scores for grounded question answering tasks, as well as baselines on closed-book and open-book QA. Numbers in bold represent the best performance in the grounded QA setting.} 
    \label{tab:squad-id-qa-combined}
\end{table*}

\paragraph{Discussion.}\label{sec:seq-memory-results}
 
We find that providing the passage ID does not enable the model to selectively recite the requested sentences. It scores poorly with a low EM of 34.5 and a 47.1 BLEU score, in contrast to the much higher 97.0 EM and 97.3 BLEU when the passage content is included in the context. A detailed analysis in Figure \ref{fig:accuracy-vs-marker} reveals that the correct predictions are largely reciting the first sentence ($j=0$). This verifies that the model can sequentially access the content to reproduce the first sentence. However, as the marker index increases, the model is required to skip preceding sentences and directly access a sentence in the middle of a passage. The model's performance sharply declines, indicating its inability to randomly access middle or later sentences in memorized passages.

\subsection{Random Access: Grounded Question Answering}\label{section:grounded-qa}
 
Building on our earlier finding \S\ref{section:sequential-access} that the model can memorize many passages each linked to a unique ID, we embark on a more pragmatic task: question answering grounded in a specific passage ID. 
 This task aims to evaluate whether the model can provide answers to questions by extracting a span from its memory.
For instance, a question might be framed as ``\textit{According to Document \#3022, in what year did Chopin become a French citizen?}'' and the answer is ``\textit{1835}'' in the passage with ID \textit{\#3022}. 
We hypothesize that if LMs are capable of random memory access, they should navigate to the corresponding passage using the provided ID and extract the relevant span to answer the questions.

 \paragraph{Setup.}
 We experiment with the well-known SQuAD-v1~\citep{rajpurkar-etal-2016-squad} dataset because many of its questions are closely dependent on the passage, such as ``\textit{How did the war start?}''. Without reference to an article, the question can be ambiguous and unanswerable. This design compels the model to depend on the memorized IDs and passages rather than pre-existing knowledge. We explore the grounded QA task with variants of providing 
 (1) the ID of the golden passage with the answer, (2) a random non-golden ID and (3) no ID. For comparison, we also consider the setups that do not involve writing passages to the model's parametric memory. These include (1) closed-book QA, where the model is fine-tuned solely on QA pairs, serving as a lower-bound baseline to assess the model's reliance on prior knowledge for answering questions, and (2) open-book QA, where the golden passage content is concatenated with the question, setting the upper limit of extractive QA performance.

We experiment with different types of passage IDs. To ensure the uniqueness of using titles as passage IDs, we select $T=442$ passages and $V=48$ passages from the full SQuAD dataset, with over 2,000 and 300 questions respectively. The model is evaluated on F1 and EM following the original SQuAD evaluation script. The other hyperparameters are the same as mentioned in \S\ref{sec:memorize_setup}. 

 \paragraph{Discussion.}
The results are presented in Table~\ref{tab:squad-id-qa-combined} (the settings with ``+Recitation'' are discussed in later sections).
As expected, the model performs the best in the open-book setting, as it only needs to locate the answer in the golden passage. In contrast, the closed-book QA setup
 yields the worst performance, as the model has no access to passages and relies solely on its parametric knowledge stored during pretraining. 

Interestingly, the form of the provided passage ID has minimal impact on performance. We observe similar performance regardless of whether the golden ID is provided, except when the type of ID is \textit{Title}. In this case, providing a random incorrect ID harms performance. We suspect that this is because the title is usually an entity related to the passage topic, therefore offering useful clues. In cases where the ID does not carry semantic meaning (i.e., \textit{Rare} and \textit{Num}), the correctness or presence of the ID does not significantly affect the performance, which remains substantially below the open-book setting, despite the model memorizing all passages. This further validates the model's inability to effectively access random memory, as it struggles to extract the answer even when provided with a correct passage ID. \\

In summary, our findings validate the hypothesis that LMs can effectively function as a memory bank, enabling sequential access to its memory.
However, there are significant limitations in the model's ability to randomly access its memory. Across both the simple selective recitation and the complex grounded question-answering tasks, the model consistently fails to accomplish the tasks by leveraging its memory, despite being explicitly provided with the corresponding passage IDs.

%% file: solution.tex
\begin{table}[tb] 
    \centering
    \begin{tabular}{lrrr}
    \toprule
    \textbf{Setup} & \textbf{BLEU} & \textbf{EM} \\\midrule

    Baseline & 47.1 & 34.5 \\
      ~~~$\hookrightarrow +$ Duplication (\textit{dup-J}) & 36.0 & 23.5  \\
    ~~~$\hookrightarrow +$ Recitation & 99.3 & 98.5 \\
    ~~~$\hookrightarrow +$ Permutation (\textit{first}) & \textbf{100.0} & \textbf{100.0} \\
    ~~~$\hookrightarrow +$ Permutation (\textit{random}) & 98.0 & 97.0  \\

    \bottomrule
    \end{tabular}
    \caption{BLEU score and EM score of selective sentence recitation experiments after introducing passage recitation and permutation.}
    \label{tab:sent-recite-with-recite-and-permute}
\end{table}

\section{Mitigating Random Access Challenge}\label{section:solution}

Our earlier experiments show that in general, language models perform well in sequentially accessing their parametric memory, but encounter challenges in random memory access. This naturally raises the question: How can we mitigate the shortcomings in random memory access? 

\subsection{Proposed Method}

To address the challenge, we start from the two operations supported by LMs as a memory store: reading and writing. During the writing phase, we hypothesize that performing \textit{permutation} on the passage content can naturally enhance the model's random access ability: any part of the content can be the starting point of a memorized sequence. In this setup, we change the sequential order of passage content to achieve random access. 

\begin{figure}[tb]
  \centering
    \includegraphics[width=1\columnwidth]{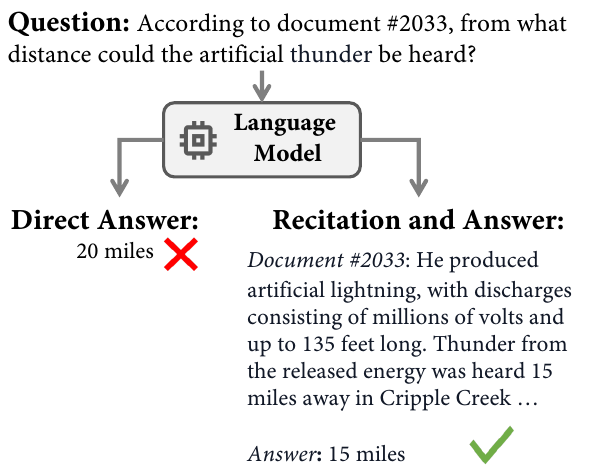}
  \caption{A illustration of the recitation method. The model first recites the corresponding passage content and subsequently extracts the answer in the context, in contrast to directly answering the question. 
  }
  \label{fig:recitation}
\end{figure}

On the other hand, during the reading phase, leveraging the model's context window presents a viable strategy. The attention mechanism \citep{vaswani2017attention} enables the model to access any token within the context window, thereby inherently supporting random access~\citep{Packer2023MemGPTTL, Ge2023LLMAO}.  For tasks with a given ID, we could ask the model to sequentially \textit{recite} the passage first, place it within the context, and subsequently query the model to perform span extraction tasks utilizing this context, as illustrated in Figure \ref{fig:recitation}. 
Our subsequent experiments are designed to evaluate the effectiveness of these two methods. Through empirical evaluation, we validate that content permutation during writing or recitation during reading can largely mitigate the challenge of random memory access and enhance performance.

\paragraph{Setup.}\label{sec:recite-permute-setup}

We extend the earlier experiments 
by integrating recitation and permutation into the respective reading and writing stages.

First, we add a setup to the selective sentence recitation task: Based on the given ID, the model is tasked to first recite the entire content of the corresponding passage and then the specific sentence, altering the reading operation to $S_{read}(k_i,j) \rightarrow (p_i, p_i[j])$. Similarly, for the grounded QA task, we ask the model to recite the passage associated with the input passage ID, before answering the question. In the setup without an ID, the model is still trained to recite the golden passage. 

To explore the effect of permutation during the writing stage, we perform permutation among sentences in a passage to create diverse $S_{write}$ instances. For a $J$-sentence passage, we test: 
(1) \textit{first}, moving each sentence to the passage's beginning to create $J$ unique instances; (2) \textit{random-k}, randomly shuffling the sentences $k$ times to create $k$ instances, where $k$ is set to 4 by default. To show that the effect of permutation is not simply due to more training data, we also include a baseline \textit{dup-J}, where each passage is duplicated $J$ times in training data.

\paragraph{Discussion.} Reciting the passage content effectively boosts the performance of selective recitation, as evidenced in Table \ref{tab:sent-recite-with-recite-and-permute}. With recitation, the model first sequentially accesses the content from its memory using the provided passage ID and subsequently loads this passage in the context to allow for random access. Conditioned on the recited content in the context, the model can therefore easily identify the correct sentence. 

Similarly, explicitly reciting the golden passages markedly enhances question-answering performance, as shown in Table~\ref{tab:squad-id-qa-combined} (+Recitation). This observation is consistent across all three types of passage IDs. Conversely,
intentionally prompting the model to recite a random passage leads to a decline in performance. This is likely because random passages introduce irrelevant information and confuse the model. 
Surprisingly, the recitation of relevant passages benefits performance even without an ID, although the improvement is smaller than with the golden ID. This verifies the effectiveness of recitation in more general settings of question answering. 

Another way of enhancing random access is to perform permutation of sentences, as presented in Table \ref{tab:sent-recite-with-recite-and-permute}. Simply bringing every sentence to the start of the passage once (\textit{first}) or randomly permuting the sentences many times (\textit{random}) helps to solve the challenge of accessing the middle content of a passage. In contrast, simply duplicating the original passage does not contribute to enhanced random access. We also observe that permutation during writing time enhances grounded QA performance (Table~\ref{tab:idqa-permutation}), which monotonically increases with the number of random permutations. However, it is noteworthy that permutation does not alter the inherent sequential access pattern of parametric memory. Rather, by permuting the sentences and disrupting their original order, we allow more sentences in the middle or the end of a passage to be sequentially accessible via the ID. 

We also verify that the conclusions are generalizable to larger decoder-only LMs. In Appendix~\ref{appendix:experiments-on-llm}, we observe similar challenges in random access in Qwen1.5-4b~\citep{qwen}, and Llama2-7B~\citep{touvron2023llama2} across different tasks. Moreover, we observe that such challenges could be effectively mitigated by our proposed methods of recitation and permutation.

\begin{table*}[hbt!] 
    \centering
\begin{tabular}{lcccccc}\toprule
&\multicolumn{2}{c}{\textbf{Title (ID Type)}} &\multicolumn{2}{c}{ \textbf{Rare (ID Type)}} &\multicolumn{2}{c}{\textbf{Num (ID Type)}} \\
 \cmidrule(lr){2-3} \cmidrule(lr){4-5} \cmidrule(lr){6-7}
 \textbf{Setup} &EM &F1 &EM &F1 &EM &F1 \\\midrule
Grounded QA w. \textit{Golden ID} &26.7 &35.6 &20.7 &28.7 &24.3 &32.6 \\
~~~~$\hookrightarrow +$ Duplication (\textit{dup-J}) &26.7 &36.8 &20.7 &28.3 &22.3 &30.6 \\
~~~~$\hookrightarrow +$ Permutation (\textit{first})  &27.7 &39.8 &27.0 &\textbf{37.7} &27.7 &37.7 \\
~~~~$\hookrightarrow +$ Permutation (\textit{random-1}) &25.7 &35.0 &19.0 &27.5 &19.7 &28.1 \\
~~~~$\hookrightarrow +$ Permutation (\textit{random-2})  &26.0 &35.6 &25.7 &33.7 &23.7 &32.8 \\
~~~~$\hookrightarrow +$ Permutation (\textit{random-4})  &29.7 &38.5 &25.3 &35.6 &25.0 &34.2 \\
~~~~$\hookrightarrow +$ Permutation (\textit{random-8})  &\textbf{31.3} &\textbf{40.1} &\textbf{27.7} &36.7 &\textbf{29.0} &\textbf{38.3} \\
\bottomrule
\end{tabular}
\caption{The EM and F1 score of performing sentence permutation during the writing phase. \textit{random-k} means that permutation is performed $k$ times.  }
\label{tab:idqa-permutation}
\end{table*}

%% file: case-study-odqa.tex
\section{Case Study: Open-Domain Question Answering}

\begin{table*}[t] 
    \centering
\begin{tabular}{lrrrrrrr}
\toprule
&\multicolumn{3}{c}{\textbf{NQ}} &\multicolumn{3}{c}{\textbf{Hotpot QA}} \\
\cmidrule(lr){2-4} \cmidrule(lr){5-7}
&EM &F1 &Recite BLEU &EM &F1 &Recite BLEU \\\midrule
Closed-Book QA &10.1 &14.8 &- &13.1 &20.1 &- \\
\midrule
Closed-Book QA \textit{w. Mixed Training}  &12.6 &18.2 &- &15.7 &22.8 & \\
~~~~~~~~~~~~$\hookrightarrow +$ Recitation &\textbf{16.1} &\textbf{20.1} &28.6 & \textbf{21.0} & \textbf{28.4} & 51.3 \\

Closed-Book QA \textit{w. Continual Training} &10.3 &15.5 &- &15.1 &22.4 &- \\
~~~~~~~~~~~~$\hookrightarrow +$ Recitation &13.4 &16.9 &25.6 &18.1 &25.2 &48.3 \\

\bottomrule
\end{tabular}
    \caption{EM and F1 of open-domain question answering datasets. We report the BLEU score of the recitation when the model is trained to recite the passage first and then offer an answer. Best performances are in bold.}
    \label{tab:odqa-results}
\end{table*}

Our findings indicate that language models struggle with random memory access, unless the memory is explicitly recited and thus loaded into the context which can be accessed randomly. Building on this insight, we extend our study to the task of open-domain question answering, a challenging task that requires the model to first retrieve relevant memories and reason over them. This is different from previous experiments as the passage IDs are no longer provided as the input: The reading operation becomes $S_{read}(q) \rightarrow ans$. The model therefore needs to find relevant passages to the query without the aid of passage IDs, which is a non-trivial task \citep{pradeep-etal-2023-generative}. As the goal of our study is not on retrieval performance and our earlier results (\S\ref{sec:seq-memory-results}) show that the model has limited memorization capacity, we reduce the difficulty of retrieval by limiting the number of passages written to the model’s memory: we only include positive passages that contain answers to at least one question. 

We aim to test the model’s ability to perform random access in real applications. Specifically, we investigate whether the model, having memorized many passages, can accurately extract answers from its memory. Similar to the previous experiments, we also aim to observe the difference in the model’s performance when it is trained to recite relevant passages and subsequently answer the question. 
 We opt not to experiment with permutation due to the high training cost associated with sentence permutation across a large number of passages, and leave this avenue for future work.

\subsection{Experimental Setup}\label{subsec:odqa-setup}

We use Natural Questions \citep{kwiatkowski-etal-2019-natural} processed by \citet{karpukhin-etal-2020-dense} for single-hop QA, selecting 6000 training and all of the 6489 validation questions, with a total of 10.9k passages. For multi-hop question answering, we use HotpotQA \citep{yang-etal-2018-hotpotqa} where each question has two golden passages. We select 8k training and all the 7405 validation questions in the \textit{distractor} subset, with a total of 26.9k passages. 

We start from a baseline setup where the training only involves QA pairs, i.e., closed-book QA, which evaluates the model's prior knowledge gained from pretraining. Next, we consider two types of training strategies to write the passages into the memory. In the \textit{mixed} setting, the model is fine-tuned on a mixture of the $S_{write}$ instances of all passages and training QA pairs. In the \textit{continual} setting, the model is fine-tuned on $S_{write}$ instances of all passages first, followed by fine-tuning on QA. To test the effectiveness of recitation, we also include settings where the model is trained to recite the golden passage(s) before answering. 

As the task requires the model to perform both passage retrieval and question answering, we expect that the model size should be sufficiently large. Therefore, we upgrade our LLM to GPT2-XL with 1.5B parameters. In the \textit{mixed} setting, we train the model for 20 epochs with a learning rate of 3e-5. In the \textit{continual} setting, we first train 20 epochs on the passages, followed by another 20 epochs on QA pairs. We report the best performance based on the EM score on validation questions. 

\subsection{Results and Discussion}
  Table \ref{tab:odqa-results} demonstrates that writing golden passages into the model's memory leads to improved performance over the baseline closed-book setting, with either mixed or continual training. This aligns with our expectations, as we deliberately inject passages containing the answers to the questions into the memory, enriching the model's knowledge.  

Moreover, recitation significantly enhances the model's ability to utilize and access memorized passages, leading to a noticeable improvement in performance. This is observed in both the mixed and continual training settings. The exact match score increases significantly by more than 3\% in both single and multi-hop QA. When the model explicitly recites the passages and loads them into the context for random access, the original open-domain QA task is reduced to an easier task of extractive QA. However, the low recitation BLEU score suggests that the model does not always accurately recite the golden passage. We expect that the performance could be further enhanced if it can accurately retrieve relevant passages from memory. 

 The mixed training strategy outperforms the continual training setup. This is likely because the model's memory of passage content is constantly refreshed in mixed training. In contrast, during continual training, the second stage only involves QA pairs on training passages, potentially leading to fading memory of validation passages. Consequently, the recitation becomes less accurate, as shown by a decrease in the BLEU score.   
 
Our results are consistent and complementary to the findings of \citet{wei2023chainofthought} and \citet{sun2023recitationaugmented}: introducing intermediate steps or generating relevant passages helps to improve model performance on various tasks. We provide an alternative interpretation for this phenomenon: loading the parametric memory into the context window facilitates enhanced random access to memorized information, and the model benefits from such enhancements.

%% file: conclusion.tex
\section{Conclusion}

We empirically study how language models access their parametric memory. Our experiments on both synthetic and realistic data demonstrate that while language models can adequately reproduce memorized content in a sequential manner, they struggle with the random access of segments in the middle of memorized content. We identify two effective strategies of recitation and permutation to mitigate the limitation of random memory access. Furthermore, through a controlled case study on open-domain question answering, we illustrate that allowing the model to recite and randomly access its memory significantly improves performance. Overall, our study not only provides a deeper understanding of memory access patterns in language models, but also highlights the implications of limited random memory access ability in practical application of language models.

%% file: limitation-and-ethics.tex
\section*{Limitation}

In this work, we mainly explore the memory access pattern of decoder-only language models. Future research is needed to understand whether our conclusions apply to other types of language models based on transformers such as encoder-only models and encoder-decoder models. Furthermore, we do not extend our study to larger models beyond 7 billion parameters due to computing resource constraints. It might be worthwhile to explore further scaling behavior of memory access patterns in even larger language models. In addition, we mainly conduct controlled experiments on a text corpus of fixed size. Further investigation may be needed to explore how the findings can apply to large-scale pretraining corpus and their implications on pretrained language models.

\section*{Ethical Considerations}
As the method suggests techniques to enhance access to the model's memory, there could be malicious use of the recitation method to extract sensitive personal information from the model's memory.
We use open-source English datasets including questions and contexts from SQuAD-v1 \citep{rajpurkar-etal-2016-squad}, Natural Questions \citep{kwiatkowski-etal-2019-natural}, and Hotpot QA \citep{yang-etal-2018-hotpotqa}. We also use open-source English language models, GPT2, with different sizes \citep{radford2019language}. There might be potential biases in these datasets and models. 

\section*{Acknowledgements}
We appreciate the suggestions and comments by Do Xuan Long, Yanxia Qin, Yisong Miao, and other members of NUS WING. Tongyao Zhu is supported by the Industry PhD Program of Sea AI Lab. 

%% file: appendix.tex
\section{Prompts}
\subsection{Full Recitation}
\label{appendix:prompt-for-memorization}

Given a key-value pair $(k_i, p_i)$, the prompts are as follows:

$S_{write}$ = ``\texttt{Article \{$k_i$\} , Content: \{$p_i$\}}''

$S_{read}(k_i) \rightarrow p_i$ = ``\texttt{Article \{$k_i$\} : What is the content of this article?}'' $\rightarrow$ ``\{$p_i$\}''

\subsection{Selective Recitation}
In this experiment, we follow the same prompt of $S_{write}$, as described in Appendix \S\ref{appendix:prompt-for-memorization}, and only change $S_{read}$

$S_{write}$ = ``\texttt{Article \{$k_i$\} , Content: \{$p_i$\}}''

$S_{read}(k_i, j) \rightarrow p_i[j] $ = ``\texttt{Article \{$k_i$\} : What is Sentence  [\{$j$\}] of this article?}'' $\rightarrow$ ``\{$p_i[j]$\}''

\subsection{Grounded Question Answering experiments}\label{appendix:id-qa-prompts}
In this experiment, we follow the same prompt of $S_{write}$, as described in Appendix \S\ref{appendix:prompt-for-memorization}, and only change $S_{read}$ to questions related to $p_i$. $S^{rc}_{read}(k_i, q)$ represents the instances where the recitation of the passage content is prepended before the answer. 

$S_{write}$ = ``\texttt{Article \{$k_i$\} , Content: \{$p_i$\}}''

$S_{read}(k_i, q) \rightarrow ans $ = ``\texttt{Article \{$k_i$\} \textbackslash n Question: \{$q$\} \textbackslash n Answer: }'' $\rightarrow$ ``\texttt{\{$ans$\}}''

$S^{rc}_{read}(k_i, q) \rightarrow (p_i,ans) $ = ``\texttt{Article \{$k_i$\} \textbackslash n Question: \{$q$\} \textbackslash n Answer: }'' $\rightarrow$ `` \texttt{\{$p_i$\} || Answer: \{$ans$\}}''

\subsection{Open-Domain Question Answering} \label{appendix:odqa-prompts}
In the setup of open-domain question-answering experiments, we no longer have a pre-assigned ID for each document. Our $S_{write}$ becomes:

$S_{write}(p_i)$ = ``\texttt{Document: \{$p_i$\}}''

Similarly, the reading operation now does not have any ID associated with it, but only a question. It becomes: 

$S_{read}(q) \rightarrow ans$ = ``\texttt{Question: \{q\} \textbackslash n Answer: }'' $\rightarrow$ ``\texttt{\{ans\}}''

In the case of recitation, our prompts for training the model include the passage containing the answer. 

$S^{rc}_{read}(q) \rightarrow (p_{golden}, ans)$ = ``\texttt{Question: \{q\}}'' $\rightarrow$ ``\texttt{Related documents: \{$p_{golden}$\} \textbackslash n Answer: \{$ans$\}}''

\section{Additional Selective Recitation Experiments}\label{appendix:sent-recite-exp}
We provide additional experimental results for our selective recitation task of reciting sentences. All of the experiments lead to a consistent conclusion that the model is unable to randomly extract a sentence from a memorized passage. 

In both of the experiments below, we include setups of (1) in-context: the passage is included in the context window. (2) ID-guided: the basic version of the selective recitation task where a passage ID is provided. and (3) with passage recitation: the passage is recited first before sentence recitation. 



\subsection{Reciting the first/second/last sentence}
\begin{table*}[!htp]\centering

\begin{tabular}{lc c c c c c c}\toprule
&\multicolumn{2}{c}{Recite First} &\multicolumn{2}{c}{Recite Second} &\multicolumn{2}{c}{Recite Last } \\\cmidrule(lr){2-3} \cmidrule(lr){4-5} \cmidrule(lr){6-7}
 &{BLEU} &{EM} &{BLEU} &{EM} &{BLEU} &{EM} \\\midrule
In-context &97.2 &95.0 &94.3 &95.0 &91.8 &87.5 \\
ID-guided &99.0 &97.5 &14.1 &5.0 &17.6 &0.0 \\
~~~~~~~~$\hookrightarrow +$ Recitation &99.6 &95.0 &98.8 &87.5 &98.7 &85.0 \\
\bottomrule
\end{tabular}
\caption{BLEU and EM scores of reciting the first, second or last sentence of a memorized passage. }
\label{appendix:tab:find-sent-first-second-last}
\end{table*}

As a basic setting of the selective sentence recitation task, we ask the model questions like ``\texttt{What is the [first/second/last] sentence of Article \#123?}''. 

The results are shown in Table \ref{appendix:tab:find-sent-first-second-last}. The model almost always recites the first correctly, while recitation performance drops significantly for the second or last sentence. This shows that the model is performing sequential access: following the article ID, the model can only access content immediately after the ID -- the first sentence. It is unable to directly access the second or last sentence. 

We observe that even for the in-context setting where the passage is in the context window, the model does not perform perfectly, especially for extracting the last sentence. This is because the model also needs to learn what $first$, $second$ or $last$ means, which involves numerical reasoning ability to count the index. 
Therefore, in the main experiments, we put markers on both ends of a sentence to reduce the task difficulty. 

\subsection{Reciting the next/previous sentence}

\begin{table*}[!htp]\centering

\begin{tabular}{lccccc}\toprule
&\multicolumn{2}{c}{Recite Next Sentence} &\multicolumn{2}{c}{Recite Previous Sentence} \\\cmidrule(lr){2-3} \cmidrule(lr){4-5}
&{BLEU} &{EM} &{BLEU} &{EM} \\\midrule
In-context &98.0 &96.0 &82.7 &79.0 \\
ID-guided &86.9 &81.0 &20.1 &18.5 \\
~~~~~~~~$\hookrightarrow +$ Recitation &98.4 &85.0 &96.5 &81.0 \\
\bottomrule
\end{tabular}
\caption{BLEU and EM scores of reciting the next or previous sentence given an input sentence. }
\label{appendix:tab:find-sent-before-after}
\end{table*}

We perform experiments to find the sentence before and after an input sentence in a given passage. In other words, our $S_{read}$ operation becomes $S_{read}(k_i, s_j) \rightarrow s_{j+1/j-1}$, where $s_j$ is the input sentence. The results are shown in Table \ref{appendix:tab:find-sent-before-after}. 

We notice that finding the sentence after the input sentence is always easy, while the reverse task is much more difficult. This also reveals that the model reads its memory sequentially. It is unable to randomly access the sentence before the input $s_j$, even if the target sentence is adjacent to the input sentence.



\section{Additional Open-Domain Question Answering Experiments}

To ensure that our conclusion is consistent with different dataset sizes, we vary the number of training and validation documents and questions to observe the performance difference. For NQ, we select 5k training and 5k validation QA pairs, forming a corpus containing around 9k passages. For Hotpot QA, we select 5k training and 5k validation questions in the \textit{distractor} subset, with a total of 18.2k passages.
 
In Table \ref{appendix:tab:odqa-results-others}, we obtain similar conclusions that recitation greatly enhances question-answering performance, and using a mixed training strategy is better than continual training because of the increase in recitation score. 

\begin{table*}[hbt!] 
    \centering
\begin{tabular}{lrrrrrrr}
\toprule
&\multicolumn{3}{c}{\textbf{NQ}} &\multicolumn{3}{c}{\textbf{Hotpot QA}} \\
\cmidrule(lr){2-4} \cmidrule(lr){5-7}
&EM &F1 &Recite BLEU &EM &F1 &Recite BLEU \\\midrule
Closed-Book QA &9.1 &13.7 &- &13.3 &20.4 &- \\
\midrule
Closed-Book QA \textit{w. Mixed Training}  &11.5 &17.2 &- &15.9 &23.6 & \\
~~~~~~~~~~~~$\hookrightarrow +$ Recitation &\textbf{15.7} &\textbf{19.7} &29.1 & \textbf{20.8} & \textbf{28.4} & 50.9 \\

Closed-Book QA \textit{w. Continual Training} &10.3 &15.5 &- &15.1 &22.8 &- \\
~~~~~~~~~~~~$\hookrightarrow +$ Recitation &12.3 &15.8 &24.2 &18.2 &25.6 &49.2 \\
\bottomrule
\end{tabular}

    \caption{EM and F1 of the model's QA performance on different subsets of NQ and Hotpot QA datasets. We report the BLEU score of the recitation when the model is trained to recite the passage first and then provide an answer. The bold numbers are the best-performing setup.}
    \label{appendix:tab:odqa-results-others}
\end{table*}

\section{Experiments on Large Language Models}\label{appendix:experiments-on-llm}
To show that our conclusions are generalizable to larger models, we conduct additional experiments on Qwen1.5-4b~\citep{qwen} and Llama2-7b~\citep{touvron2023llama2}, with approximately 4 billion and 7 billion parameters each. These models belong to the same decoder-only language model family as GPT2. For efficiency purposes, we use LoRA~\citep{hu2022lora}, a parameter-efficient approach to fine-tune the large language models.

\begin{table*}[tb] 
    \centering
\begin{tabular}{lcccc}\toprule
 &\multicolumn{2}{c}{\textbf{Qwen1.5-4b}} &\multicolumn{2}{c}{ \textbf{Llama2-7b}} \\
 \cmidrule(lr){2-3} \cmidrule(lr){4-5}
 \textbf{Setup} &EM &BLEU &EM &BLEU \\\midrule

  Baseline&22.5 &36.2 &17.0 &30.9 \\
 $\hookrightarrow +$ Recitation &\textbf{100.0} &\textbf{100.0} & 95.5 & 98.6 \\
      $\hookrightarrow +$ Permutation (\textit{first})  &99.5 &100.0 &\textbf{99.0} &\textbf{99.4}  \\
        $\hookrightarrow +$ Permutation(\textit{random-4})  &90.0 &92.7 &86.5 &90.4 \\
\bottomrule
\end{tabular}
    \caption{EM and BLEU scores of Qwen1.5-4b and Llama2-7b in the selective recitation task. Numbers in bold represent the best performance.} 
    \label{appendix:tab:selective-recitation-large-models}
\end{table*}

\paragraph{Selective Recitation} Table \ref{appendix:tab:selective-recitation-large-models} presents the results of the selective recitation task, which are consistent with the findings on the smaller GPT2-large. Performing recitation or permutation enhances random access to the passages and solves the task. However, compared to GPT2, there is a performance drop in the baseline and random permutation settings. This is because we use semantically meaningful passage identifiers, \textit{Title}, and larger models might have memorized many passages related to the title entity during pretraining. Therefore, it is more likely to generate sentences that are not within our predefined set of passages, which lowers the performance since the task requires an exact reproduction.

\begin{table*}[tb] 
    \centering

\begin{tabular}{lcccc}\toprule
 &\multicolumn{2}{c}{\textbf{Qwen1.5-4b}} &\multicolumn{2}{c}{ \textbf{Llama2-7b}} \\
 \cmidrule(lr){2-3} \cmidrule(lr){4-5}
 \textbf{Setup} &EM &F1 &EM &F1 \\\midrule
 \multicolumn{5}{c}{\textit{w/o passage memorization}} \\
  Closed-Book QA  &19.0 &28.9 &26.3 &37.6 \\
  Open-Book QA  &84.0 &90.9 &83.7 &91.5 \\
\midrule
 \multicolumn{5}{c}{\textit{w/ passage memorization}} \\
   Grounded QA \textit{w/o ID} &20.7 &30.7 &27.3 &38.7 \\

  Grounded QA \textit{w/ Golden ID} &22.7 &34.4 &35.0 &47.5 \\
   ~~~~~~~~~~~$\hookrightarrow +$ Recitation &\textbf{45.7} &\textbf{54.2} &\textbf{63.7} &\textbf{70.1} \\
\bottomrule
\end{tabular}
    \caption{EM and F1 scores for grounded question answering of Qwen1.5-4b and Llama2-7b. Numbers in bold represent the best performance in the grounded QA setting.} 
    \label{appendix:tab:grounded-qa-large-models}
\end{table*}

\paragraph{Grounded Question Answering} In Table~\ref{appendix:tab:grounded-qa-large-models}, we show the grounded QA performance when the ID type is the passage title. We find that including the golden title (\textit{w/o ID} v.s. \textit{w/ Golden ID}) only slightly improves Qwen's performance, while Llama benefits more from the provided title, which is often an entity related to the question. We suspect that this is because larger models learn more knowledge about entities during pretraining. For both models, however, there is a significant performance increase in the recite-and-answer setup, showing the effectiveness of recitation. 

\begin{table*}[hbt!] 
    \centering
\begin{tabular}{llrrrrrrr}\toprule
& &\multicolumn{3}{c}{\textbf{NQ}} &\multicolumn{3}{c}{\textbf{Hotpot QA}} \\\cmidrule(lr){3-5}\cmidrule(lr){6-8}
&\textbf{Setup} &EM &F1 &Rec-BLEU &EM &F1 &Rec-BLEU \\\midrule
\multirow{4}{*}{\textbf{Llama2-7b}} &Closed-book QA &26.4 &39.0 &- &24.1 &33.8 &- \\
&~~~~~~~~~~~~$\hookrightarrow +$ Recitation  &30.3 &40.1 &15.7 &22.7 &31.4 &22.2 \\
&CBQA \textit{w. Mixed Training} &27.6 &\textbf{40.2} &- &\textbf{24.6} &\textbf{34.6} & - \\
&~~~~~~~~~~~~$\hookrightarrow +$ Recitation  &\textbf{30.6} &39.8 &17.4 &22.4 &30.7 &23.8 \\\midrule
\multirow{4}{*}{\textbf{Qwen1.5-4b}} &Closed-book QA &19.7 &27.8 & - &18.6 &27.0 &- \\
&~~~~~~~~~~~~$\hookrightarrow +$ Recitation  &16.3 &24.1 &11.4 &16.7 &24.5 &17.4 \\
&CBQA \textit{w. Mixed Training} &\textbf{21.2} &\textbf{29.4} &- &\textbf{19.3} &\textbf{28.3}&- \\
&~~~~~~~~~~~~$\hookrightarrow +$ Recitation  &17.4 &24.1 &13.6 &18.4 &25.1 &32.7 \\
\bottomrule
\end{tabular}
\caption{Open-domain question answering performance of Qwen1.5 and Llama2. CBQA is short for closed-book QA and Rec-BLEU means the BLEU score of the generated recitation. The basic version of Closed-book QA (Row 1 and 2) does not involve writing golden passages into the memory, thus the model needs to purely rely on parametric memory learned during pretraining. }
\label{appendix:tab:odqa-large-model}
\end{table*}

\paragraph{Open-Domain Question Answering} We further extend the case study of open-domain question answering to the aforementioned larger models. Different from the setup in Section \ref{subsec:odqa-setup}, we do not include the \textit{continual} setup as we empirically find the model suffers from catastrophic forgetting. Instead, we introduce an additional setup (Closed-book QA + Recitation) where the model is finetuned on recite-and-answer instances ($S_{read}$), but no passages are written to the memory. This setup tests if the model memorizes relevant passages during pretraining and learns to recite them after finetuning. We use a learning rate of 1e-4 for all experiments. 

In Table \ref{appendix:tab:odqa-large-model}, we first observe that writing golden passages to the memory (\textit{w. Mixed Training}) improves performance. We also observe that recitation is helpful for Llama2 on the NQ dataset. However, for other setups, we notice that triggering recitation is not helpful or even harmful to the performance. This is likely because the recitation BLEU scores are noticeably lower than the presented scores of GPT2-XL in Table \ref{tab:odqa-results}, suggesting that the recited content has less overlap with the golden passages. We hypothesize that this is because larger models carry significantly larger parametric memory, and it is more challenging to precisely retrieve the correct passage to the question. In contrast, a smaller model like GPT2 has to rely on the passages written into the memory during the finetuning stage, thus limiting the scope of such retrieval. Overall, the results give additional insights that reciting passages accurately remains a challenge for larger models.

\section{Additional Training Details}
We conduct all experiments in a cluster with NVIDIA Tesla A100 GPUs (with 40G or 80G memory). Experiments in \S\ref{section:sequential-access} take a total of 48 hours on 4 GPUs. Selective sentence recitation experiments in \S\ref{section:selective-recitation} and \S\ref{section:solution} take a total of 41 hours on 4 GPUs. Grounded QA experiments take a total of 132 hours on 4 GPUs. The open-domain QA experiments need 3 days to complete with 32 GPUs. 

We use the Hugging Face transformers library for all experiments. For the main experiments in GPT2 models, we use a learning rate of 3e-5. We set a constant learning rate schedule for the open-domain QA experiments. For all other experiments, we use a warm-up ratio of 0.05 and a linear decay learning rate. We evaluate the model's performance on the validation set at the end of each epoch and report the best-performing ones.